\documentclass[superscriptaddress,prb,11pt]{revtex4-1}

\usepackage{amsmath}    
\usepackage{graphicx}   
\usepackage{verbatim}   
\usepackage{color}      
\usepackage{subfigure}  
\usepackage{hyperref}   
\begin{document}

\title{Making up for the deficit in a marathon run}

\author{Iztok Fister Jr.}
\affiliation{University of Maribor, Faculty of Electrical Engineering and Computer Science, Smetanova 17, 2000 Maribor, Slovenia; Email: iztok.fister1@um.si}
\author{Du\v{s}an Fister}
\affiliation{University of Maribor, Faculty of Mechanical Engineering, Smetanova 17, 2000 Maribor, Slovenia}
\author{Suash Deb}
\affiliation{IT \& Educational Consultant, Ranchi, Jharkhand, India}
\affiliation{Distinguished Professorial Associate, Decision Sciences and Modelling Program, Victoria University, Melbourne, Australia}
\author{Uro\v{s} Mlakar}
\affiliation{University of Maribor, Faculty of Electrical Engineering and Computer Science, Smetanova 17, 2000 Maribor, Slovenia}
\author{Janez Brest}
\affiliation{University of Maribor, Faculty of Electrical Engineering and Computer Science, Smetanova 17, 2000 Maribor, Slovenia}
\author{Iztok Fister}
\affiliation{University of Maribor, Faculty of Electrical Engineering and Computer Science, Smetanova 17, 2000 Maribor, Slovenia}

\date{10 February 2017}

\begin{abstract}
To predict the final result of an athlete in a marathon run thoroughly is the eternal desire of each trainer. Usually, the achieved result is weaker than the predicted one due to the objective (e.g., environmental conditions) as well as subjective factors (e.g., athlete's malaise). Therefore, making up for the deficit between predicted and achieved results is the main ingredient of the analysis performed by trainers after the competition. In the analysis, they search for parts of a marathon course where the athlete lost time. This paper proposes an automatic making up for the deficit by using a Differential Evolution algorithm. In this case study, the results that were obtained by a wearable sports-watch by an athlete in a real marathon are analyzed. The first experiments with Differential Evolution show the possibility of using this method in the future. 
\end{abstract}

\maketitle

\section{Introduction}
Running a marathon is a very challenging task for every athlete. Good and efficient preparation for a marathon is a very complex process that must be performed many time before the competition. The marathon is not just an ordinary running competition, but it is an epic distance having roots in ancient Greece. The marathon name was coined by the legend of soldier Philippides, who ran from the Battle of Marathon to Athens in order to announce the victory of the united Greeks against the Persian force. 

Nowadays, most of the big cities organize marathon competitions annually that attract a lot of people. Some athletes just want to finish a marathon, while others have higher goals. In line with this, the amateur athletes are satisfied with an achievement of around four hours, while more competitive amateur athletes try to beat the so-called magic line, the three hours. Although this magic result was set many years ago, running a marathon race sub three hours is still considered as a good result for a well-trained amateur athletes. However, those who would like to reach the result of sub three hours not only need the proper sports training, but also a lot of knowledge about a race course. Additionally, awareness of characteristics about an athlete's body is a crucial step in the evolution of runners, i.e., when to go faster, when to eat, when to slow down, how to pace on hills are some of the most significant questions that runners ask themselves when planning their running pace. 
 
Making up for the deficit time in a marathon run is a very demanding task~\cite{Fister2015CiSport} that comes into play when the final results as predicted by sports trainers are slower than the achieved one by only a few seconds. Thus, more experienced athletes also calculate the pace for each kilometer of a race many weeks before the race start in order to predict their final achievement. However, the achieved result can distinguish from the predicted, because of objective, as well as subjective factors. The objective factors represent environmental conditions such as weather, temperature, wind position and even humidity, while the subjective factors refer to the athlete's psycho-physical conditions such as feelings on the race day. Nowadays, predicting the final result is a little bit easier using the modern mobile technology. In this group, we can count professional sports-watches that consist also of Global Positioning System (GPS) sensors~\cite{Fister2013WidespreadTriathlons}, heart rate measuring sensors, cadence meters, etc. Additionally, there is a comprehensive software support and some intelligent systems have also been arisen recently. 

In this paper, we present a novel solution for making up for the deficit marathon run that is based on the data collected from sports-watches worn by the athlete during the run. The primary aim of this study is to help athletes who did not achieve the predicted final results by a few seconds and show them on which parts of the predefined course they could improve the intermediate pace so that the predicted results could be reached. This solution is implemented using Differential Evolution (DE)~\cite{storn1997differential}, which calculates the improvement of the intermediate pace, based on the configuration of the running course. These improvements are considered as constraints in the DE and they base on the traditional rule of the marathon run, i.e., more seconds can be caught up when the running course is flat. The secondary aim of this approach is also to help athletes in their analysis of the already performed marathon race. In our case study, we focus on the situation where an athlete tries to achieve the magic marathon time of three hours.

The paper is structured in the remainder as follows. In Section~\ref{Sec2}, making up for the deficit in a marathon run is formulated as a Constraint Satisfaction Problem (CSP). Section~\ref{de:chap} is devoted to Differential Evolution. Section~\ref{Sec3} illustrates the case study, in which the case of an athlete in the Three hearts marathon in 2012 is taken into consideration. Experiments are presented in the second part of Section~\ref{Sec3}. The paper is concluded with Section~\ref{Sec5}, where the work done is reviewed and directions are outlined for further research.

\section{Problem formulation} \label{Sec2}
Making up for the deficit time can be formulated as a CSP~\cite{eiben2003introduction}, where the solution is specified as a vector $\mathbf{p}=(p_1,\ldots ,p_n)^T$ of $n$ problem variables representing the intermediate pace deficit for each of the observed kilometer. Additionally, a variable bound constraint is attached to each problem variable that limits the proper values of each variable to the interval $p_i\in [p^{(L)}_i,p^{(U)}_i]$, where $p^{(L)}_i$ and $p^{(U)}_i$ are lower and upper bounds, respectively. In the case of a marathon race, where the length of the whole course equals 42.195~km, the length of each vector is, consequently, $n=43$ elements. Then, the problem is defined in generalized form as follows:
\begin{equation}
\begin{aligned} 
\max \quad & f(\mathbf{p})=\sum_{i=1}^{n}{p_i}\leq T_a-T_p, & \\ 
\text{subject to} \quad & p^{(L)}_i \leq p_i \leq p^{(U)}_i, & i=1,\ldots,n;\\
\end{aligned}
\normalsize
\label{eq:moop}
\end{equation}
\noindent where $T_a-T_p$ denotes the deficit time that must be made up and the relation $T_a\geq T_p$ is valid. Thus, the following inequality constraints must be satisfied:
\begin{equation}
\sum_{i=1}^{n}{p_i^{(L)}}\geq T_a-T_p, 
\label{eq:obj}
\end{equation} 
\indent which ensures that the problem is solvable. Although the upper bounds can be set as negation of the corresponding lower bounds (i.e., $p^{(U)}_i=-p^{(L)}_i$ in general, here the upper bounds are fixed to zero (i.e., $p^{(U)}_i=0$). This mean that only negative values of intermediate pace deficits are allowed in our study. 

 
The lower bound values $p^{(L)}_i$ are calculated from the configuration of the course according to the following assumptions:
\begin{equation}
\mathit{Alt}_i=\left\{ \begin{array} {lll} 
=0, & \text{flat}, & \mapsto -2~\text{sec}, \\
<0, & \text{downhill}, & \mapsto -4~\text{sec}, \\
>0, & \text{uphill}, & \mapsto 0~\text{sec}, \\
\end{array} \right.
\label{eq:alt}
\end{equation}
\noindent where $\mathit{Alt}_i$ is an average altitude obtained in the appropriate kilometer. The speculation behind the assumptions is to speed up, when the course is flat or downward and to retain the speed when running upward. The quantitative values for estimation of deficit time is also taken from the real marathon practice, where it holds that the deficit time can be a maximum of 4 seconds in one kilometer when the course is flat.

\section{Differential Evolution}
\label{de:chap}
Differential Evolution (DE) is an Evolutionary Algorithm appropriate for continuous and combinatorial optimization that was introduced by Storn and Price in 1995~\cite{storn1997differential,das2016recent}. DE is a population-based algorithm~\cite{fister2013brief} that consists of $\mathit{Np}$ real-coded vectors representing the candidate solutions, as follows:
\begin{equation}
\label{eq:de_repr}
 \mathbf{x}_{i}^{(t)}=(x_{i,1}^{(t)}, \ldots ,x_{i,n}^{(t)}),\ \ \ \textnormal{for}\ i=1, \ldots, \mathit{Np},
\end{equation}
\noindent where each element of the solution is in the interval $x_{i,1}^{(t)}\in [x_{i}^{(L)},x_{i}^{(U)}]$, and $x_{i}^{(L)}$ and $x_{i}^{(U)}$ denote the lower and upper bounds of the $i$-th variable, respectively. 

The variation operator in DE supports a differential mutation and a differential crossover. In particular, the differential mutation selects two solutions randomly and adds a scaled difference between these to the third solution. This mutation can be expressed as follows:
\begin{equation}
\label{eq:de_mut}
 \mathbf{u}_{i}^{(t)}=\mathbf{x}_{r1}^{(t)}+F \cdotp (\mathbf{x}_{r2}^{(t)}-\mathbf{x}_{r3}^{(t)}),\ \ \ \textnormal{for}\ i=1, \ldots, \mathit{Np},
\end{equation}
\noindent where $F$ denotes the scaling factor as a positive real number that scales the rate of modification while $r1,\ r2,\ r3$ are randomly selected values in the interval $[1 \ldots \mathit{Np}]$. Note that, typically, the interval $F \in [0.1,1.0]$ is used in the DE community.

As a differential crossover, uniform crossover is employed by the DE, where the trial vector is built from parameter values copied from two different solutions. Mathematically, this crossover can be expressed as follows:
\begin{equation}
\label{eq:de_xover}
 w_{i,j}^{(t+1)}=\begin{cases}
          u_{i,j}^{(t)} & \textnormal{rand}_{j}(0,1) \leq \mathit{CR} \vee j=j_{\mathit{rand}}, \\
		  x_{i,j}^{(t)} & \text{otherwise},
        \end{cases}
\end{equation}
\noindent where $\mathit{CR} \in [0.0,1.0]$ controls the fraction of parameters that are copied to the trial solution. Note, the relation $j=j_{rand}$ ensures that the trial vector is different from the original solution $\mathbf{x}_{i}^{(t)}$.

A differential selection is, in fact, a generalized one-to-one selection that can be expressed mathematically as follows:
\begin{equation}
\label{eq:de_sel}
 \mathbf{x}_{i}^{(t+1)}=\begin{cases}
          \mathbf{w}_{i}^{(t)} &\text{if } f(\mathbf{w}_{i}^{(t)}) \leq f(\mathbf{x}_{i}^{(t)}), \\
		  \mathbf{x}_{i}^{(t)} &\text{otherwise}\,.
        \end{cases}
\end{equation}
In a technical sense, crossover and mutation can be performed in several ways in Differential Evolution. Therefore, a specific notation is used to describe the varieties of these methods (also strategies) generally. For example, 'DE/rand/ 1/bin' denotes that the base vector is selected randomly, 1 vector difference is added to it, and the number of modified parameters in the mutant vector follows binomial distribution. 

\subsection{Making up for the deficit time using DE}
Modifying the original DE to solve a problem of making up for the deficit time in a marathon run (let us call a new variant marathon DE or simply mDE) is relatively simple, because DE implements the variable bounds implicitly. Therefore, the main problem remains how to set the appropriate variable bounds for elements of solution vectors. The answer to this question is given when intermediate pace of an athlete is taken into consideration. The intermediate paces are, nowadays obtained using the mobile sports-watches worn by the marathon runner during the run. 

There are four phases of modifying the original DE for making up for the deficit time in a marathon run, as follows:
\begin{itemize}
\item defining the objective
\item preparing the bounds for problem variables
\item mapping the solution from decision space to problem space
\item evaluating the solution in problem space
\end{itemize}
Indeed, defining the objective refers to the expression $T_a-T_p$ in Eq.~(\ref{eq:obj}), where we define the total deficit of time that must be made up. Preparing the bounds for problem variables $p_{i,j}$ bases on tracked data from sports-watches referring to the information about altitude as overcome by the marathon runner in each kilometer (Eq.~(\ref{eq:alt})). According to these data, the appropriate boundary values are calculated according to Eq.~(\ref{eq:alt}). The mapping of solution from decision space $\mathbf{x}^{(t)}_i$ to its counterpart in the problem space $\mathbf{p}^{(t)}_i$ is straightforward and obeys the following equation:
\begin{equation}
p^{(t)}_{i,j}=\left \lceil \frac{p^{(U)}_{j}-p^{(L)}_{j}}{x^{(U)}_{j}-x^{(L)}_{j}}\cdot x^{(t)}_{i,j}\cdot 10 \right \rceil \cdot 10^{-1},
\end{equation}
\noindent where multiplying by $10^{-1}$ ensures that the given result in seconds is extended with the most significant digit of milliseconds after decimal point. Finally, the fitness function presented in Eq.~(\ref{eq:moop}) is used in our study.

In order to simplify solving of the problem, the last element in vector $\mathbf{p}^{(t)}_{i}$ that does not represent the whole kilometer is excluded from the optimization, i.e., it is taken as an uphill part, although it can be downhill in practice as is in our case. 

\section{Case study: The Three hearts \\marathon 2012} \label{Sec3}
\begin{figure}[htb]
\includegraphics[width=0.5\textwidth]{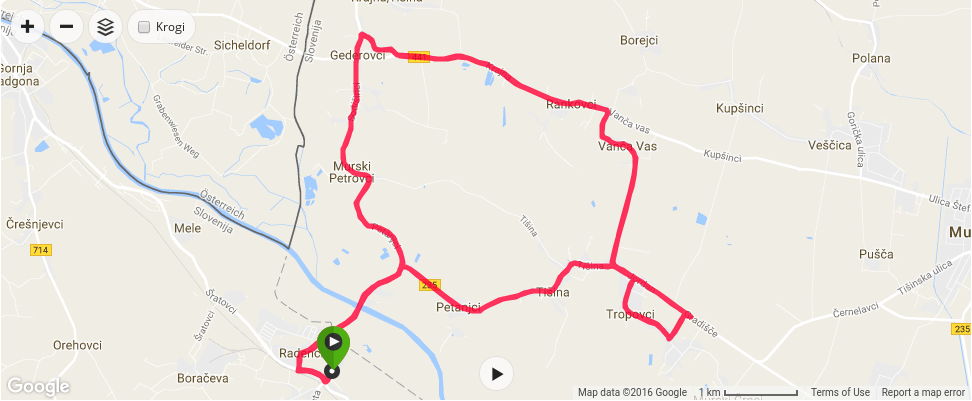}
\caption{Three Hearts marathon course in 2012.}
\label{RadenciCourse}
\end{figure}

\begin{figure}[htb]
\includegraphics[width=0.5\textwidth]{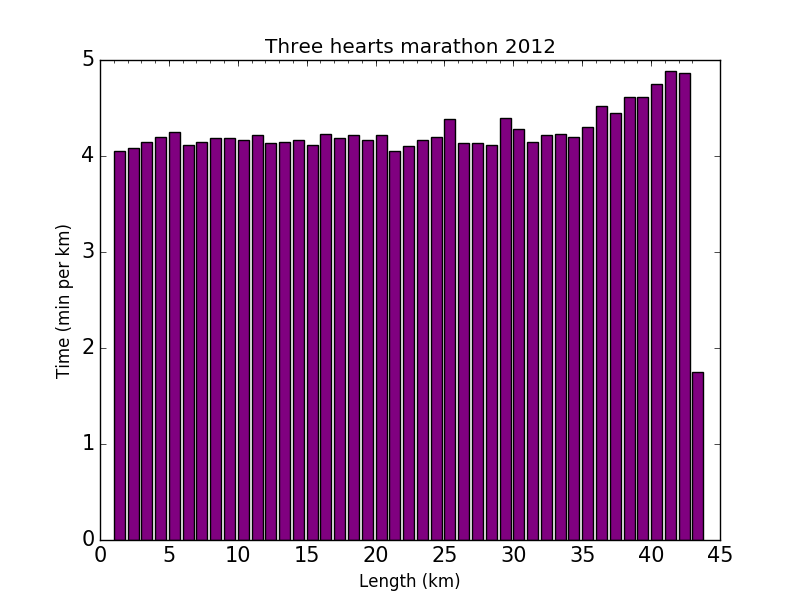}
\caption{Average pace for each kilometer.}
\label{RadenciPace}
\end{figure}

The Three Hearts marathon has been organized annually in the city of Radenci (Slovenia) since 1981. During the years, the organizers have changed the marathon course slightly. In 2012, the marathon course consisted of two 21.1 kilometer long laps as can be seen in Figure~\ref{RadenciCourse}. 
The goal of an athlete in our study was to run the marathon sub 3:00 hours. However, to run the marathon in this time, the intermediate pace should be on average 4:15 minutes per kilometer (min/km) (Figure~\ref{RadenciPace}) or better (less than 4:15). 
The athlete's plan was to run the first kilometers a little bit faster than the others. Although the course was flat on average, he did not focus enough on some ascents with a slight uphill that were large enough to lose a power, especially in bad weather conditions. After 30 km, the time was still good enough for achieving the result under 3:00 hours, but the last kilometers were pretty hard to overcome because of a formerly broken nail, abrasions from the heat and many blisters. Therefore, the athlete was one minute slower at the finish than planned. 

\begin{figure}[htb]
\label{altitude}
\includegraphics[width=0.5\textwidth]{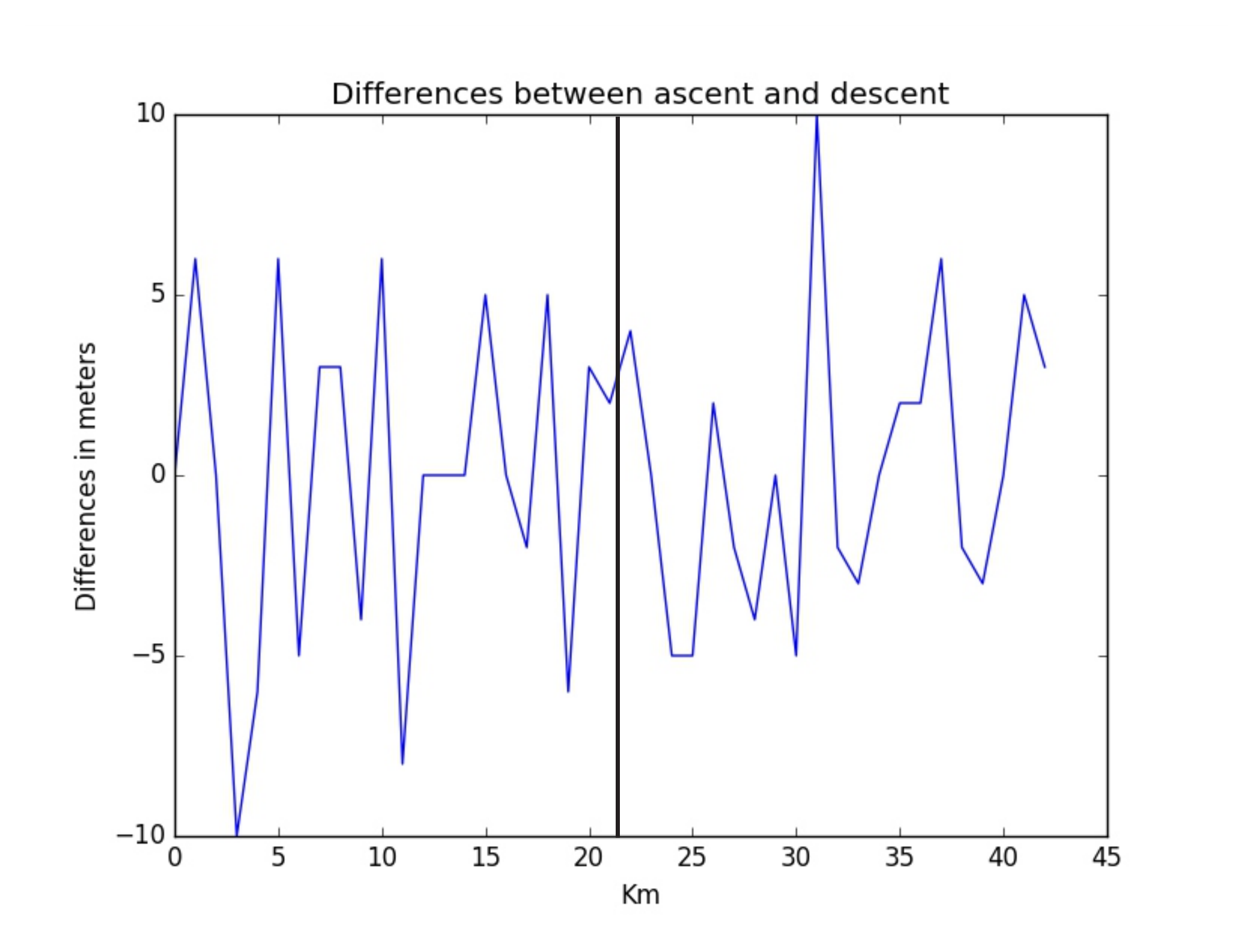}
\caption{Ascents and descents on the course of Three Hearts marathon at 2012.}
\end{figure}

The proposed approach for making up for the deficit time works as follows. The objective is expressed as: 
\begin{equation*}
3\!:\!01\!:\!09.4-3\!:\!00\!:\!00.0=1\!:\!09.4~\min \quad \approx 70~\text{sec}.
\end{equation*}
An altitude graph of the marathon course, from which problem variable bounds were prepared, is presented in Figure~3.

The graph in Figure~3 presents differences of altitudes in meters according to the run distances in kilometers that is divided into two parts denoting 21.1~km by a vertical line. Although both parts must be the same in theory, this assumption does not hold in practice, because GPS technology, used by sports-watches, is far from ideal for now. However, the tracked data are accurate enough for the purposes of our study.

The results of an analysis of altitude data are as follows. The GPS receiver detected 4 flat parts, 22 uphill`s and 17 downhills. The last 195 meters were downward. Essentially, the difference between the number of uphill`s and downhills is a consequence of errors in the GPS receiver. On the other hand, there were no serious uphill`s and downhills on the marathon course as can be seen from the maximum and minimum differences of the altitude (i.e., $\pm 10$ meters). In fact, the difference between average ascent and average descent values at the end of each kilometer higher than $\pm 1$ meter is declared as uphill or downhill in the study.

Bounds for problem variables $\mathbf{p}^{(t)}_i$ were prepared based on the analysis of the altitude data according to Eq.~(\ref{eq:alt}). The results of this phase are illustrated in Table~\ref{rezultadores} under columns "Lower bounds" and "Upper bounds". The sum of the lower bounds of all problem variables is $\sum^{n}_{j=1}{p^{(L)}_{i,j}}=82$, whereby a condition of the problem solvability in Eq.~(\ref{eq:obj}) was satisfied.

The last two phases, i.e., mapping and evaluating the solution, were performed using the mDE. For our experiments, mDE was coded in Python programming language and the 'DE/rand/1/bin' strategy was employed. The input of the algorithm were data obtained from the Three Hearts marathon. The dimension of the problem was $n=43$, population size was set to $\mathit{Np}=100$, scaling factor $F=0.5$ and crossover rate $\mathit{CR}=0.9$. Lower and upper bounds were set in the interval $x^{(t)}_{i,j}\in [0,1]$. The algorithm terminated when either fitness function fulfilled the objective and, thus, all constraints were satisfied 100 times, or the number of fitness function evaluations exceeded 80,000. Only one run of the mDE was performed in this study.

The obtained results of the mDE algorithm are presented in Table~\ref{rezultadores} that consists of six columns. The first column counts kilometers during the marathon run, while the second illustrates actual intermediate pace, as already presented in Figure~\ref{RadenciPace}. The third column depicts the predicted intermediate pace obtained after decrementing the actual intermediate pace by the intermediate pace deficit found in the first run that is presented in column four. The last values are obtained after conducting the mDE algorithm. As already mentioned, the last two columns present lower and upper bounds. 

\begin{table*}[htb]
\caption{Results of making up for the deficit in the Three Hearts marathon}
\tiny
\begin{center}
\begin{tabular}{|c|c|c|c|c|c|}
\hline
Distance & Actual pace (sec)  &  Predicted pace (sec)  &  Difference  & Lower bounds & Upper bounds \\ 
$[\text{km}]$ & [min/km] & [min/km] & [sec] & [sec] & [sec] \\  \hline
1 & 04:03.80 & 04:03.80 & 0 & 0 & 0 \\ 
2 & 04:05.40 & 04:03.60 & 1.8 & 2 & 0 \\ 
3 & 04:09.70 & 04:05.90 & 3.8 & 4 & 0 \\ 
4 & 04:12.20 & 04:08.50 & 3.7 & 4 & 0 \\ 
5 & 04:15.00 & 04:15.00 & 0 & 0 & 0 \\ 
6 & 04:07.70 & 04:04.30 & 3.4 & 4 & 0 \\ 
7 & 04:09.70 & 04:09.70 & 0 & 0 & 0 \\ 
8 & 04:11.20 & 04:11.20 & 0 & 0 & 0 \\ 
9 & 04:11.00 & 04:07.80 & 3.2 & 4 & 0 \\ 
10 & 04:10.30 & 04:10.30 & 0 & 0 & 0 \\  \hline
11 & 04:13.80 & 04:10.00 & 3.8 & 4 & 0 \\ 
12 & 04:08.80 & 04:07.00 & 1.8 & 2 & 0 \\ 
13 & 04:09.10 & 04:07.30 & 1.8 & 2 & 0 \\ 
14 & 04:10.70 & 04:09.70 & 1 & 2 & 0 \\ 
15 & 04:07.90 & 04:07.90 & 0 & 0 & 0 \\ 
16 & 04:14.40 & 04:12.70 & 1.7 & 2 & 0 \\ 
17 & 04:11.20 & 04:07.30 & 3.9 & 4 & 0 \\ 
18 & 04:13.00 & 04:13.00 & 0 & 0 & 0 \\ 
19 & 04:10.00 & 04:06.40 & 3.6 & 4 & 0 \\ 
20 & 04:13.60 & 04:13.60 & 0 & 0 & 0 \\  \hline
21 & 04:03.00 & 04:03.00 & 0 & 0 & 0 \\ 
22 & 04:06.40 & 04:06.40 & 0 & 0 & 0 \\ 
23 & 04:10.00 & 04:08.30 & 1.7 & 2 & 0 \\ 
24 & 04:12.00 & 04:08.50 & 3.5 & 4 & 0 \\ 
25 & 04:23.80 & 04:21.20 & 2.6 & 4 & 0 \\ 
26 & 04:08.40 & 04:08.40 & 0 & 0 & 0 \\ 
27 & 04:08.00 & 04:04.30 & 3.7 & 4 & 0 \\ 
28 & 04:07.60 & 04:03.80 & 3.8 & 4 & 0 \\ 
29 & 04:24.40 & 04:22.80 & 1.6 & 2 & 0 \\ 
30 & 04:17.00 & 04:14.40 & 2.6 & 4 & 0 \\  \hline
31 & 04:09.30 & 04:09.30 & 0 & 0 & 0 \\ 
32 & 04:13.70 & 04:10.30 & 3.4 & 4 & 0 \\ 
33 & 04:14.70 & 04:10.80 & 3.9 & 4 & 0 \\ 
34 & 04:12.60 & 04:11.00 & 1.6 & 2 & 0 \\ 
35 & 04:18.60 & 04:18.60 & 0 & 0 & 0 \\ 
36 & 04:31.90 & 04:31.90 & 0 & 0 & 0 \\ 
37 & 04:27.80 & 04:27.80 & 0 & 0 & 0 \\ 
38 & 04:37.70 & 04:33.90 & 3.8 & 4 & 0 \\ 
39 & 04:37.40 & 04:34.60 & 2.8 & 4 & 0 \\ 
40 & 04:45.20 & 04:43.70 & 1.5 & 2 & 0 \\  \hline
41 & 04:53.70 & 04:53.70 & 0 & 0 & 0 \\ 
42 & 04:52.40 & 04:52.40 & 0 & 0 & 0 \\ 
42.195 & 01:45.30 & 01:45.30 & 0 & 0 & 0 \\  \hline
Total: & 03:01:09.40 & 02:59:59.40 & 70 & 82 & 0 \\  \hline
\end{tabular}
\end{center}
\label{rezultadores}
\normalsize
\end{table*}

\subsection{Discussion}
At a first glance, the conducted experiments are in contrast with the principles of stochastic population-based optimization, where we are interested in the mean values obtained after the number of runs. Strictly speaking, this is true, when the results are obtained from the aforementioned point of view. However, the aim of our study was twofold: On the one hand, to show that the mDE algorithm can be employed for making up the deficit in a marathon race, while on the other hand, to prove that there are many solutions of this problem. Both assumptions were justified by finding numerous feasible solutions in just one run. Here, the question arises "Could we say, which of the solutions found is really optimal?". However, the answer to the question must be answered by professionals, i.e., sports trainers and coaches together with athletes.

\section{Conclusion} \label{Sec5}
This paper reports the first successful application of the mDE for making up for the deficit in a marathon run based on the history data obtained by a sports-tracker on the corresponding marathon course. Making up for the deficit in a marathon run is a complex task that is usually performed by professional trainers and even athletes themselves. We defined this problem as a constraint problem and solved it by the mDE. Preliminary experiments shown that proposed approach may be used in the real-world. 

However, there are still many open directions for further research in this problem, e.g., to determine the most appropriate solution for a specific athlete from a huge set of solutions. In line with this, context dependent information need to be accumulated, on which basis such a solution can be proposed by the mDE algorithm automatically. However, extending input data with a heart rate monitor might also increase the prediction possibilities of the proposed algorithm.

\end{document}